\documentclass[11pt]{article}

\usepackage{microtype} 
\usepackage{booktabs}  
\usepackage{url}  
\usepackage{graphicx}
\usepackage{float}

\usepackage{amsmath}
\usepackage{amsthm}

\usepackage[final, hidesupplement]{automl}

\usepackage{times}
\usepackage{epsfig}
\usepackage{graphicx}
\usepackage{amsmath}
\usepackage{bbm}

\usepackage{slashbox}
\usepackage{tabularx,ragged2e}
\usepackage{spreadtab}
\usepackage{adjustbox}
\usepackage{booktabs}
\usepackage{multirow}
\usepackage{rotating}
\usepackage{color, colortbl}
\usepackage{subcaption}
\usepackage{tcolorbox}
\usepackage{color,soul}

\usepackage{cleveref}

\graphicspath{{./figures/}}

\DeclareGraphicsExtensions{.jpeg,.png,.eps,.pdf}

\newcolumntype{a}{>{\columncolor{Gray}}c}

\newcommand{\partitle}[1]{\bigbreak\noindent\textbf{#1}}

\makeatletter
\newcommand*{\rom}[1]{\expandafter\@slowromancap\romannumeral #1@}
\makeatother

\usepackage{mathtools}

\usepackage{pifont}

\newlist{todolist}{itemize}{2}
\setlist[todolist]{label=$\square$}
\usepackage{pifont}

\makeatletter
\newcommand*\bigcdot{\mathpalette\bigcdot@{.5}}
\newcommand*\bigcdot@[2]{\mathbin{\vcenter{\hbox{\scalebox{#2}{$\m@th#1\bullet$}}}}}
\makeatother

\definecolor{applegreen}{rgb}{0.55,0.71,0.0}
\definecolor{babyblue}{rgb}{0.54,0.81,0.94}
\definecolor{azure}{rgb}{0.0,0.5,1.0}
\definecolor{budgreen}{rgb}{0.48,0.71,0.38}
\definecolor{amaranthpurple}{rgb}{0.67,0.15,0.31}

\newcommand{\ie}{\textit{i}.\textit{e}., }

\crefformat{section}{\S#2#1#3} 
\crefformat{subsection}{\S#2#1#3}
\crefformat{subsubsection}{\S#2#1#3}

\usepackage{booktabs,arydshln}

\makeatletter
\def\adl@drawiv#1#2#3{%
        \hskip.5\tabcolsep
        \xleaders#3{#2.5\@tempdimb #1{1}#2.5\@tempdimb}%
                #2\z@ plus1fil minus1fil\relax
        \hskip.5\tabcolsep}
\newcommand{\cdashlinelr}[1]{%
  \noalign{\vskip\aboverulesep
           \global\let\@dashdrawstore\adl@draw
           \global\let\adl@draw\adl@drawiv}
  \cdashline{#1}
  \noalign{\global\let\adl@draw\@dashdrawstore
           \vskip\belowrulesep}}
\makeatother

\definecolor{lightcyan}{rgb}{0.88,1,1}
\definecolor{applegreen}{rgb}{0.55, 0.71, 0.0}
\definecolor{aqua}{rgb}{0.0, 1.0, 1.0}
\definecolor{beaublue}{rgb}{0.74, 0.83, 0.9}
\definecolor{blond}{rgb}{0.98, 0.94, 0.75}
\definecolor{caribbeangreen}{rgb}{0.0, 0.8, 0.6}
\definecolor{classicrose}{rgb}{0.98, 0.8, 0.91}
\definecolor{darkseagreen}{rgb}{0.56, 0.74, 0.56}
\definecolor{lightgreen}{rgb}{0.56, 0.93, 0.56}
\definecolor{mediumaquamarine}{rgb}{0.4, 0.8, 0.67}
\definecolor{babypink}{rgb}{0.96, 0.76, 0.76}
\definecolor{cambridgeblue}{rgb}{0.64, 0.76, 0.68}
\definecolor{celadon}{rgb}{0.67, 0.88, 0.69}
\definecolor{etonblue}{rgb}{0.80, 0.92, 0.64}

\newcommand{\automl}{AutoML}

\usepackage{amsmath}
\usepackage{bm}
\usepackage[utf8]{inputenc} 
\usepackage[T1]{fontenc}    
\usepackage{hyperref}       
\usepackage{url}            
\usepackage{nicefrac}       
\usepackage{microtype}      
\usepackage{xcolor}         

\def\1{\bm{1}}

\DeclareMathAlphabet{\mathsfit}{\encodingdefault}{\sfdefault}{m}{sl}
\SetMathAlphabet{\mathsfit}{bold}{\encodingdefault}{\sfdefault}{bx}{n}

\usepackage{tikz}

\usepackage{hyperref}
\usepackage{url}
\usepackage{multirow}
\usepackage{enumitem}

\usepackage{amsthm}
\usepackage{mathtools}
\theoremstyle{plain}

\usepackage{varwidth}
\usepackage{enumitem}
\usepackage{tikz}
\usetikzlibrary{positioning}
\usetikzlibrary{shapes}
\usetikzlibrary{fit}
\usetikzlibrary{calc}

\usepackage{listings}
\usepackage{xcolor}
\usepackage{courier}

\definecolor{codegreen}{rgb}{0,0.6,0}
\definecolor{codegray}{rgb}{0.5,0.5,0.5}
\definecolor{codeblack}{rgb}{0.,0.,0.}
\definecolor{codepurple}{rgb}{0.58,0,0.82}
\definecolor{backcolour}{rgb}{0.95,0.95,0.92}

\lstdefinestyle{mystyle}{
    backgroundcolor=\color{backcolour},   
    commentstyle=\color{codegreen},
    keywordstyle=\color{codeblack},
    numberstyle=\tiny\color{codegray},
    stringstyle=\color{codepurple},
    basicstyle=\ttfamily\footnotesize,
    breakatwhitespace=false,         
    breaklines=true,                 
    captionpos=b,                    
    keepspaces=true,                 
    numbers=left,                    
    numbersep=5pt,                  
    showspaces=false,                
    showstringspaces=false,
    showtabs=false,                  
    tabsize=2,
    aboveskip=0pt,
    belowskip=-3pt
}

\newcommand{\Quote}[2]{
}

\usepackage{pifont}

\usepackage{natbib}

\usepackage{natbib}

\title{What Can \automl{} Do For Continual Learning?}

 \author[1]{\nameemail{Mert Kilickaya}{kilickayamert@gmail.com}}
 \author[1]{\nameemail{Joaquin Vanschoren}{j.vanschoren@tue.nl}}

\affil[1]{Automated Machine Learning Group, Eindhoven University of Technology}

\hypersetup{%
  pdfauthor={}, 
  pdftitle={},
  pdfsubject={},
  pdfkeywords={}
}

\begin{document}

\maketitle

\begin{abstract}


This position paper outlines the potential of \automl{} for incremental (continual) learning to encourage more research in this direction. Incremental learning involves incorporating new data from a stream of tasks and distributions to learn enhanced deep representations and adapt better to new tasks. However, a significant limitation of incremental learners is that most current techniques freeze the backbone architecture, hyperparameters, and the order \& structure of the learning tasks throughout the learning and adaptation process. We strongly believe that \automl{} offers promising solutions to address these limitations, enabling incremental learning to adapt to more diverse real-world tasks. Therefore, instead of directly proposing a new method, this paper takes a step back by posing the question: "What can \automl{} do for incremental learning?" We outline three key areas of research that can contribute to making incremental learners more dynamic, highlighting concrete opportunities to apply AutoML methods in novel ways as well as entirely new challenges for AutoML research.
\end{abstract}

\begin{figure*}[!h]
    \centering
\includegraphics[width=0.8\textwidth]{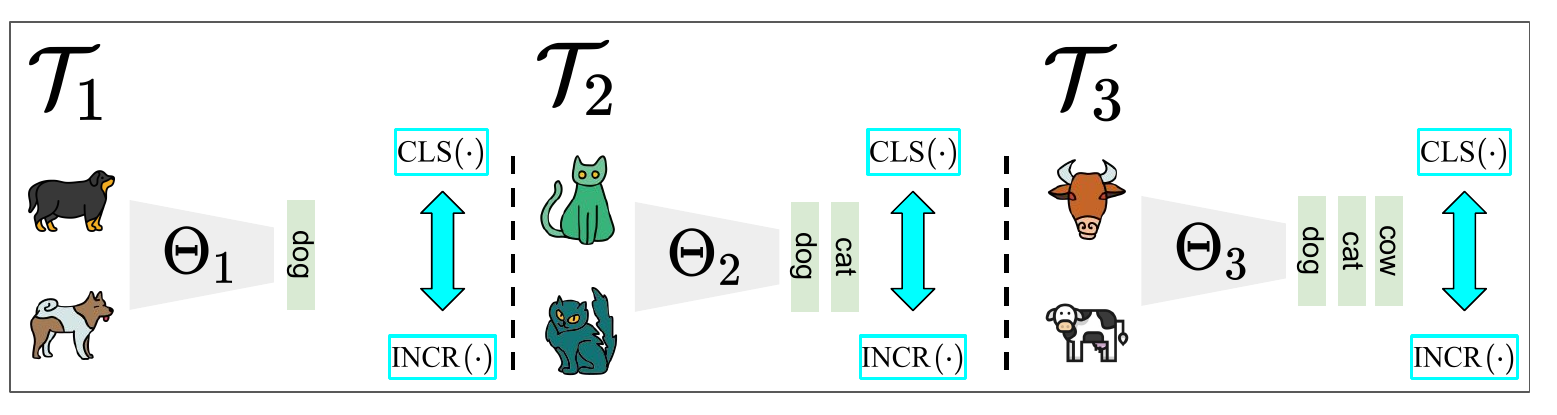}
\caption{An example of a class-incremental learning pipeline. Incremental learning consists of three main blocks: \textit{(i) The learning task sequence} - here: classifying dogs, cats, and cows occurs in three separate tasks, \textit{(ii) The learning backbone (architecture)}, parameterized by $\Theta$, which might need to be adapted throughout the whole learning process, and \textit{(iii) The learning objective}, simultaneously optimizing adaptation to the new problem, e.g. cross-entropy loss for class-incremental tasks ($\text{CLS}$), and minimizing incremental losses ($\text{INCR}$) such as `forgetting' concepts learned in previous tasks. We believe that \automl{} has the potential to transform all three of these components: defining a better task curriculum, optimizing the model architecture, and tuning the learning objective for each new task.} 
\label{fig:teaser}

\end{figure*}

\section{Motivation}

Incremental (continual) learning updates deep representations with new data distributions~\citep{masana2020class}, as illustrated in Figure~\ref{fig:teaser}. An incremental learning pipeline consists of three main components. Firstly, the learner receives a set of learning tasks, each with a corresponding dataset. Secondly, a backbone model (architecture), such as a ResNet~\citep{he2016deep}, is trained on top of the data. Lastly, a learning objective is computed to update the model parameters. This objective involves both adaptation and incremental learning losses. We want the model to adapt to new tasks (e.g. learn new classes) as fast as possible. At the same time, the incremental learning loss prevents \textit{catastrophic forgetting} of previous knowledge, sometimes referred to as anti-forgetting loss. To prevent forgetting, researchers either regularize network weights across learning sessions~\citep{kirkpatrick2017overcoming}, or store and replay a subset of the previous training data~\citep{lopez2017gradient}. 

Regardless of the methodology, all three components share a common limiting assumption: staticity. The order and the structure (\ie the categories of incremental learning tasks) of learning tasks, the architecture of the backbone learner, and the hyperparameters of the learning objective remain fixed throughout the incremental learning process. We strongly believe that \automl{}~\citep{he2021automl} has the potential to make incremental learners more dynamic. Therefore, in this paper, we shift our focus from engineering new models and instead explore the various ways in which \automl{} can be applied in the context of incremental learning.

\begin{figure}[h]
    \centering
\includegraphics[width=\textwidth]{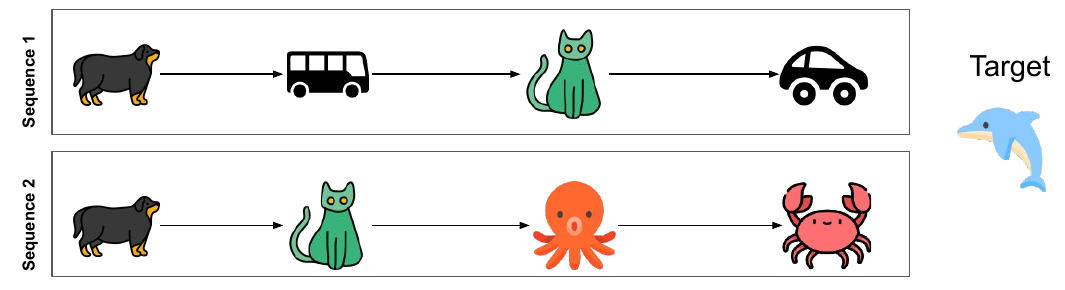}


\caption{Consider a target learning task focused on \textit{dolphins}. Amongst the two learning sequences, we believe the second one is a more viable choice, as it includes generic animals, followed by sea animals prior to dolphins. This sequential progression may enhance the learner's ability to grasp the unique features and nuances associated with the dolphin category.}

\label{fig:order}

\end{figure}

\section{\automl{} Can Search for Incremental Learning Tasks}




\partitle{Background.} In incremental learning, the learner is presented with a sequence of $t$ learning tasks:
\begin{align}
\mathcal{T}_{1:t} = (\mathcal{T}_{1}, \mathcal{T}_{2}, ...,\mathcal{T}_{t})
\end{align}
Each learning task $\mathcal{T}_{\tau}$ is accompanied by a corresponding dataset $\mathcal{D}_{\tau} = {(\mathbf{x}_{i,\tau}, \mathbf{y}_{i,\tau})^{n_{\tau}}}$, where $\tau$ indexes the current learning task, $\mathbf{x}_{i,\tau}$ represents the input, and $\mathbf{y}_{i,\tau}$ denotes the corresponding label. Each input pair $(\mathbf{x}_{i,\tau}, \mathbf{y}_{i,\tau}) \in \mathcal{X}_{\tau} \times \mathcal{Y}_{\tau}$ is sampled from an unknown distribution. It is important to note that the learning tasks are mutually exclusive, meaning that the label sets $\mathcal{Y}_{\tau-1}$ and $\mathcal{Y}_{\tau}$ have no overlapping categories. In other words, the learner encounters a completely new set of categories during each learning increment.




\partitle{Limitation.} A limitation of the current learning task arrangement lies in the order and the structure (\ie the type of the learning tasks) of tasks presented to the learner. These tasks are typically sampled randomly, provided in an arbitrary order, which does not align with the natural learning progression observed in lifelong human learners~\citep{khan2011humans}. In human learning, we typically start with the fundamentals of a topic and gradually increase the level of difficulty.

Consider Figure~\ref{fig:order}. Our ultimate objective is to learn about dolphins, and we are presented with two possible learning sequences. The first sequence combines animals and vehicles, while the second sequence starts with animals and then progresses to sea animals before focusing on dolphins. Among these options, we argue that following sequence 2, which incorporates a more transferable progression of knowledge relevant to our target learning goal, is more suitable, highlighting the need for automatic design of learning sequences. 



\partitle{What Can \automl{} Do?} \automl{} can optimize task sequence design via Curriculum Learning~\citep{soviany2022curriculum}. In curriculum learning, the goal is to find the optimal order of the data that will improve the performance on the downstream target task (\ie dolphin classification). In case of incremental learning, curriculum learning can be repurposed to identify the order as well as the structure of the tasks, instead of the training data itself. 



\partitle{Preliminary Work.} In their pioneering work, Bell and Lawrence~\citep{bell2022effect} first show that the order in which incremental learning tasks are presented to the learner highly influences the results. And then, they propose an automatic technique, that given the space of learning tasks, outputs the optimal order which will yield the highest incremental learning accuracy and reduce forgetting. However, the contribution is mostly executed on simplistic, synthetic datasets, and only considers the order and not structure, leaving room for real-life experiments combining both.

\begin{figure}[h]
    \centering
\includegraphics[width=0.5\textwidth]{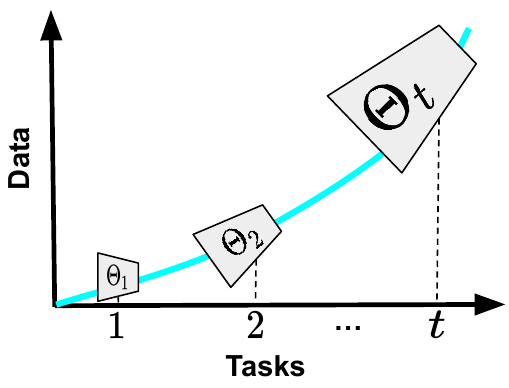}
\caption{During incremental learning, as the model observes more tasks, the volume of the observed data may increase exponentially. In that manner, it is advantageous to grow the model (\ie deeper, wider) to better fit the observed data.}
\label{fig:nas}
\end{figure}

\section{\automl{} Can Search for Incremental Learning Architectures}


\partitle{Background.} When a new learning task arrives, a deep neural network $f_{\Theta}$ with parameters $\Theta$ is trained to embed the input to the output space: $f_{\Theta}^{\tau}: \mathcal{X}_{\tau}\rightarrow \mathcal{Y}_{\tau}$, where $f_{\Theta}^{\tau}$ represents the parameters of the learner during the learning increment $\tau$. As a result, at the end of the learning, we have $t$ different states of the learner: 
\begin{align}
[f_{\Theta}^{1}, f_{\Theta}^{2}, \ldots, f_{\Theta}^{t}] 
\end{align}
\noindent each starting from the previously trained model weights. 





\partitle{Limitation.} The limitation of the backbone is the fixed capacity\footnote{Here, we note the literature of architectural incremental learning~\citep{wang2023task,rusu2016progressive,serra2018overcoming,li2019learn,veniat2020efficient}. These works still fix the architecture of the backbone learner prior to incremental learning, and then learn to allocate subsets of the fixed network to incoming learning increments. To that end, the capacity of the backbone learner remains constant.} of the learner through all learning sessions. Incremental learners are implemented with deep neural networks such as a ResNet-$18$~\citep{he2016deep}, and the model weights are updated with the learning task data. However, the architecture remains the same, even though the learning tasks are getting more complex: More categories to distinguish across, and more data to fit. It is well known that bigger models perform better with larger-scale data~\citep{ramasesh2022effect,niu2021adaxpert}, and that architecture matters in continual learning problems~\cite{DBLP:journals/corr/abs-2202-00275}.

Ideally, we should be able to automatically adjust the size of the backbone to fit the needs of the learning task(s). Consider Figure~\ref{fig:nas}. Here, incremental learning starts with a relatively smaller network. As the size of the observed data and categories grows, so does the capacity of the incremental learner. 


\partitle{What Can \automl{} Do?} \automl{} can optimize the learning backbone with respect to the learning tasks, with the help of Neural Architecture Search (NAS)~\citep{white2023neural}. A particular challenge here is to be extremely efficient, (re)designing neural architectures that achieve maximum performance while minimizing computation (\ie FLOPS). Specific NAS concepts such as compound scaling \citep{DBLP:journals/corr/abs-1905-11946} or meta-learning \citep{DBLP:journals/corr/abs-1911-11090, Gastel} can play a significant role here. Moreover, while traditional NAS methods optimize downstream task loss, e.g. cross-entropy loss for classification, incremental learning requires this to be coupled with across-task anti-forgetting objectives, and hence include incremental learning methods such as weight and data regularization or episodic memory.





\partitle{Preliminary Work.} The pioneering work of~\citep{gao2022efficient} explores the intersection of incremental learning and NAS. More specifically, the authors propose a Reinforcement Learning agent that, given a new learning task, finds the optimal set of neurons to discard and add to the current learner that maximizes accuracy, and minimizes forgetfulness. \cite{DBLP:journals/corr/abs-2202-00275} empirically analyzed the impact of various architectural components and provide best practices and recommendations to improve continual learning.

\begin{figure}[h]
    \centering
\includegraphics[width=\textwidth]{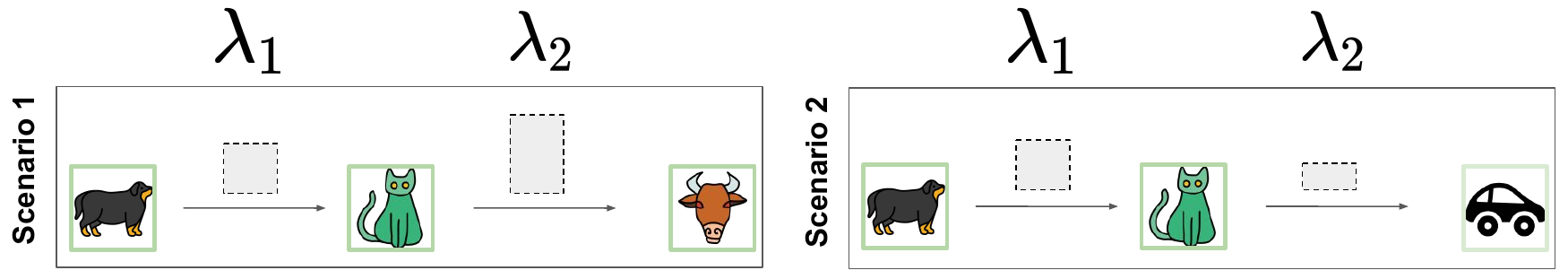}
\caption{We depict two scenarios, where the learner observes either cows or cars, after learning tasks of dogs and cats. In the first scenario, the model already is exposed to semantically similar (animal) categories. As a result, the incremental learner may not need to learn everything from scratch, hence downgrading the importance of the classification loss. Contrary to that, to learn about a completely distinct concept of car, it may be viable to anti-forget less and learn more.}
\label{fig:hyper}
\end{figure}

\section{\automl{} Can Search for Incremental Learning Hyperparameter}

\partitle{Background.} An incremental learner optimizes the following objective: 
\begin{align}
    \label{eq:main}
    \text{CLS}(\cdot) + \lambda\cdot\text{INCR}(\cdot)
\end{align}
\noindent where $\text{CLS}(\cdot)$ is the standard Cross-Entropy loss to correctly classify the current learning task. Here, $\text{INCR}(\cdot)$ is the incremental learning loss that tries to retain the past information, either via regularizing the neural network weights to change smoothly across learning sessions~\citep{lopez2017gradient} or storing and replaying a subset of the previous training data for replay~\citep{kirkpatrick2017overcoming,chaudhry2018riemannian,zenke2017continual,lee2017overcoming}. Here, $\lambda$ controls the contribution of learning new information via classification loss \textit{vs.} retaining past information via incremental learning loss. 


\partitle{Limitation.} The limitation of the hyperparameter is that $\lambda$ is constant throughout all learning increments. This assumes that regardless of the set of learning tasks, the model will always require a fixed amount of classification and anti-forgetting objectives.

Such an assumption is unrealistic, consider Figure~\ref{fig:hyper}. Here, we depict two scenarios, where the learner observes either cows or cars, after learning tasks of dogs and cats. In the first scenario, the model already is exposed to semantically similar (animal) categories. As a result, incremental learner may not need to learn everything from scratch, hence downgrading the importance of the classification loss. Contrary to that, to learn about a completely distinct concept of car, it may be viable to anti-forget less and learn more. 





\partitle{What Can \automl{} Do?} \automl{} can predict the magnitude of anti-forgetting per-task via Hyperparameter Optimization (HPO)~\citep{yu2020hyper}. In HPO, the goal is to predict the hyperparameters automatically, with little to no human involvement. Two prominent techniques include either Bayesian optimization~\citep{snoek2012practical} or gradient-based techniques~\citep{baydin2018automatic}. 



\partitle{Preliminary Work.} We identify two concurrent works to perform auto-tuning of the $\lambda$ parameter. Firstly, ~\citep{liu2023online} proposes a reinforcement learning (RL) approach, where an oracle RL agent predicts $\lambda$, trains the underlying incremental learner, and gets rewarded for performing well on the current and previous learning tasks. 

Secondly, ~\citep{gok2023adaptive} follows a Bayesian-Optimization approach, where they treat the incremental learner as a black box. The authors predict the best $\lambda$ value per-task, with the guidance of a validation loss over current and previously observed data. Both papers present a significant improvement over the conventional, fixed-hyperparameter counterparts, motivating further research into automatic tuning of $\lambda$.

\section{Conclusion}



\partitle{Summary.} This paper discusses the potential broader impact of \automl{} on incremental learning, which would enable incremental learning techniques to tackle a much wider range of real-world streams of tasks that require fast adaptation. These currently lie beyond the capabilities of both incremental learning and AutoML techniques. 

We first identify a limiting assumption in incremental learners, which pertains to their static nature in terms of choosing the sequence of learning tasks, architecture, and hyperparameters. To overcome these limitations, we explore the question, "What can \automl{} do for incremental learning?" We propose making incremental learners dynamic by incorporating automated curriculum learning to design streams of learning tasks, neural architecture search to design task-specific architectures, and optimization techniques to tune hyperparameters (including those used in loss functions) on-the-fly.


\partitle{Limitations.} We acknowledge three limitations in our work. Firstly, we focus on identifying \textit{what} \automl{} tools can be applied to improve incremental learning, rather than providing a detailed explanation of \textit{how} they should be applied. Therefore, we include pioneering preliminary work to inspire future research in enhancing incremental learning with \automl{}. 

Secondly, although significant progress has been made, most \automl{} techniques require substantial computational resources. Applying these techniques directly to incremental learning would necessitate re-running the process for each incremental step, resulting in time-consuming computations. Hence, future work should explore methods for scaling sub-linearly with the number of learning tasks.

Lastly, in this paper, we discuss the potential benefit of \automl{} for incremental learning. An equally interesting question would be the benefit of incremental learning for \automl{}, to use incremental learning paradigm to make \automl{} techniques more efficient. 

%
%
%
%
%

{\small
\bibliographystyle{apalike}
\bibliography{egbib}
}

\end{document}